# An optimization-based IMU/Lidar/Camera Co-calibration method

1st Lanhua Hou
*Graduate School of Engineering*
*(The University of Tokyo)*
*School of Instrument Science*
*(Southeast University)*
Tokyo, Japan
houlanhua@g.ecc.u-tokyo.ac.jp

2nd Xiaosu Xu*
*School of Instrument Science and Engineering*
*(Southeast University)*
Nanjing, China
xxs@seu.edu.cn

3rd Takuma Ito*
*Graduate School of Engineering*
*(The University of Tokyo)*
Tokyo, Japan
takumaito@g.ecc.u-tokyo.ac.jp

4th Yiqing Yao
*School of Instrument Science and Engineering*
*(Southeast University)*
Nanjing, China
yucia@sina.com

*Abstract*—Recently, multi-sensors fusion has achieved significant progress in the field of automobility to improve navigation and position performance. As the prerequisite of the fusion algorithm, the demand for the extrinsic calibration of multi-sensors is growing. To calculate the extrinsic parameter, many researches have been dedicated to the two-step method, which integrates the respective calibration in pairs. It is inefficient and incompact because of losing sight of the constrain of all sensors. With regard to remove this burden, an optimization-based IMU/Lidar/Camera co-calibration method is proposed in the paper. Firstly, the IMU/camera and IMU/lidar online calibrations are conducted, respectively. Then, the corner and surface feature points in the chessboard are associated with the coarse result and the camera/lidar constraint is constructed. Finally, construct the co-calibration optimization to refine all extrinsic parameters. We evaluate the performance of the proposed scheme in simulation and the result demonstrates that our proposed method outperforms the two-step method.

## I. INTRODUCTION

Intelligent mobility plays a more and more important role in the field of autonomous driving, disaster relief, exploration, life service, and so on. To establish a high-precision and robust perception and localization system, multi-sensor fusion technology has been widely applied in intelligent mobility[1]. Especially, camera, Lidar, and IMU occupy very important positions with their complementing advantages. Camera can provide rich, colorful but unstable semantic information and Lidar gives accurate, robust but sparse depth information, which is resistant to illumination. IMU is high-autonomy but its error accumulates over time. To achieve reciprocal advantages, the information from different sensors is delicately fused to construct an accurate and robust fusion system[2]. Now, IMU/Camera/Lidar integrated system is increasingly becoming the focus of multi-sensor systems for intelligent mobility.

To fuse sensors precisely, information from all sensors should be unified in one coordinate system by calculating the relative 6-Dimension pose of each sensor in the procedure of calibration. Calibration is the prerequisite of location and perception for intelligent mobility and has been widely and deeply researched recently. Especially, plentiful researches focus on the calibration of pairwise sensors, such as IMU/camera, IMU/Lidar and Lidar/camera[3]. But there are fewer researches focusing on more than two sensors, especially for three different sensors. The usual calibration method for more than two sensors is based on the simple combination of pairwise sensor, ignoring the constrain crossing all sensors, which will lead to unstable and inaccurate result in complex environments. In order to solve this problem, a more accurate co-calibration IMU/Camera/Lidar algorithm is proposed in this paper. The contributions of this paper are as follows:

1) Improve accuracy of IMU/Camera/Lidar calibration by the mutual constraints of IMU/camera, IMU/Lidar and Lidar/camera;

2) Provide initial value for Lidar/camera calibration by IMU/camera and IMU/Lidar sub-calibration system;

3) Lidar/camera calibration was carried out based on the line and surface features of the chessboard to improve the observability of the calibration system;

4) Provide a real-time IMU/Camera/Lidar co-calibration method.

The remainder of this paper is organized as follows. Section II gives a brief review of related work. Section III presents the detail of the proposed algorithm. Section IV provides the experimental results to show the precise of the proposed method. Finally, conclusions and open issues are discussed in Section V.

## II. RELATIVE WORK

### A. IMU/Camera Calibration

Thanks to the advancement of IMU and camera, IMU/camera integrated navigation system has been widely applied in autonomous mobility. And the calibration of the two sensors has been researched as it affects the accuracy of the integrated system. IMU/camera calibration can be divided into offline calibration and online calibration. Offline calibration needs to be carried out in advance but it is high-precision. Faraz M et al. proposed a chessboard-based IMU/camera calibration method with the time correlations of the IMU and uncertainties computing[4]. Rehder, J et al. published kalibr, an open-source calibration tool for IMU/camera, which is based on a chessboard and can simultaneously calibrate intrinsic and extrinsic parameters of IMU and camera[5].

Online calibration is convenient, robust and fast. Nevertheless, the accuracy is lower than the offline calibration because of the IMU's error accumulation. To deal with the external disturbances in the application of Visual-Inertial Odometry(VIO), Xiao et al. proposed an online calibration monitor to recalibrate the extrinsic parameter as it changes[6].

*B. IMU/Lidar Calibration*

Lidar, as a 3D perception sensor, is the pioneer to solve the 3D simultaneous, location and mapping(SLAM) problem. To introduce the IMU to improve the performance of Lidar-based SLAM, the calibration of IMU/Lidar is of great importance. Similar to IMU/camera calibration, IMU/Lidar calibration can be divided into offline and online calibration. Offline calibration needs to be carried out in advance, which is accurate but inconvenient. Lv et al. developed a continuous-time-based IMU/Lidar calibration method to deal with the Lidar distortion in the high maneuver condition, and then introduced the observability-aware modules of segmentation and degeneration to improve the accuracy and robustness of this system[7][8].

Online calibration is convenient and fast but the accuracy is lower because of IMU's error accumulation. Cedric Le Gentil et al. researched the unsampled pre-integrated of IMU information to deal with the motion distortion in the IMU/Lidar online calibration system[9]. Liu et al. proposed a target-less-based multi-feature online calibration method, which detects the point/sphere, line/cylinder and plane features for Lidar registration[10]. Zuo et al. published the LIC-Fusion system, an MSCKF-based Lidar/IMU/camera online calibration and location system[11]. Then, they creatively introduced a sliding window filter and outlier rejection to the system to improve the accuracy and robustness[12].

*C. Camera/Lidar Calibration*

In the field of autonomous mobility, camera and Lidar based 3D perception has been extensively applied subsequent to the wide use of Lidar and camera. And plenty of research strived on the calibration of camera and Lidar, which greatly affects the precision and robustness of the 3D perception. Camera/Lidar calibration can be divided into the target-based and target-less methods. Target-based calibration needs to prepare the targets to generate the relation formulation of the two sensors for calibration. Verma S et al. researched a line and plane based camera/Lidar calibration with a checkboard and demonstrated that parallel boundaries substantially degraded the calibration system[13]. Kümmerle et al. constructed an estimation problem to calculate the intrinsic and extrinsic parameters simultaneously with a spherical calibration target[14]. Kim et al. converted the 2D-3D registration to 3D-3D registration calibration problem with a planar chessboard[15]. Xie et al. developed a novel circular-holes-based chessboard and proposed a new detective algorithm for Lidar and camera calibration. Choi et al. developed a 3D calibration target to calibrate the thermal infrared camera and Lidar[16].

The target-less method is convenient and useful. Pandey et al. utilized the surface intensities of Lidar and camera to construct the target-less estimation problem for camera and Lidar. But the unknown initial value results in slower convergence. Zhen et al. proposed a camera/Lidar-based structure from motion problem with the online IMU/Lidar calibration simultaneously[17]. Nagy et al. converted 2D-3D calibration problem to 3D-3D registration problem by structure from motion of camera[18]. Yuan et al. proposed a target-less-based Lidar/camera calibration method in the natural environment by matching the natural edge features of Lidar and camera[19].

*D. Multi-sensors*

With the omnidirectional localization and perception requirements of autonomous driving, more and more sensors are introduced, such as Lidar, camera, radar, sonar, wheel odometry, IMU and so on. Generally, with the increase of the sensors, the performance of the multi-sensor system becomes better. The performance of the multi-sensor system extremely depends on the calibration. To apply the multi-sensor system, it is necessary to calibrate them simultaneously. There are many researches about multi-camera, multi-Lidar, multi-IMU, multi-camera and Lidar, multi-Lidar and camera, IMU and multi-camera while rare researches emphasize the simultaneous calibration of more than three kinds of sensors[20][21][22] [23] [24] [25]. Domhof et al. developed an IMU/Lidar/radar co-calibration system and demonstrated the robustness of this system[26]. Pentek et al. presents a target-less Lidar-GNSS/IMU-camera joint calibration system for the unmanned aerial vehicle by estimating Lidar-GNSS/IMU and GNSS/IMU-camera extrinsic in sequence[27]. The only method about IMU/Lidar/camera calibration based on Kalman filter, which is imprecise to deal with the linear problem in SLAM, is mentioned in the preprint paper[28]. All of these researches demonstrate that the co-calibration of all sensors is more accurate and robust than the divide calibration. Therefore, an optimization-based IMU/Lidar/camera calibration is imperative for autonomous mobility.

In this paper, an optimization-based joint calibration algorithm with the geometric constraint of three pairs of sensors is proposed. At the same time, the calibration results of IMU/camera and IMU/Lidar are utilized as the initial value of camera/Lidar to facilitate the detection of a chessboard. Then the line and plane feature points are introduced to improve the observability of the calibration system. Finally, the co-calibration problem is proposed and solved.

## III. METHOD

This paper developed a high-precision joint calibration algorithm and the overall framework of the proposed algorithm is shown in Fig.1. There are mainly four steps to deal with the information from three sensors: IMU/camera online sub-calibration system, IMU/Lidar online sub-calibration system, Lidar/camera line/plane association, and IMU/Lidar/camera co-calibration system.

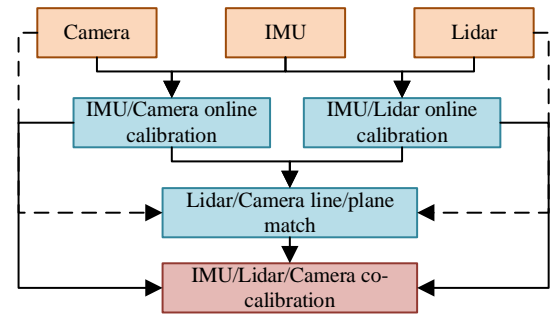

Fig 1. The system structure of the proposed method.

As shown in Fig.2, at first, the extrinsic parameter of IMU/camera is introduced as a state vector to VIO system, which is inspired by VINS-MONO. Similarly, the extrinsic parameter of IMU/Lidar is introduced as a state vector to Lidar inertial odometry system and the Lidar deskew is performed based on the constant velocity assumption but not the IMU because of the unknown transformation matrix. Third, based on the results of the IMU/Lidar and IMU/camera sub-calibration systems, the initial transformation matrix from camera to Lidar is calculated. Then, the line and surface feature points in the calibration board from the two sensors are extracted and associated. Finally, the overall IMU/Lidar/camera system is constructed based on the estimation results of the IMU/Lidar and IMU/camera subsystems and the line and surface features to construct the joint optimization of the three sensors. The feature association based on the initial estimation improves the accuracy of line-surface feature matching, and the point-surface feature association establishes calibration constraints for each sensor, which improves the accuracy of the calibration system.

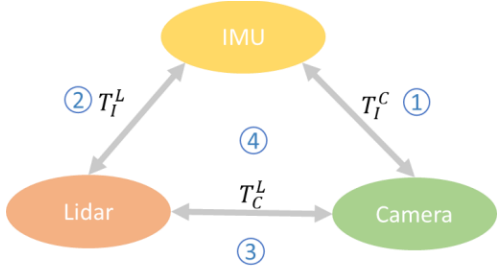

Fig 2. The system structure of the proposed method.

### A. IMU/Camera Calibration

To ensure the fast convergence of the online calibration of IMU/camera, the hand-eye calibration method is applied to roughly estimate the rotation of IMU and camera:

$$T_{C_0}^{I_0} T_{C_n}^{C_0} T_{I_n}^{C_n} = T_{I_n}^{I_0} \quad (1)$$

Where, $T_{C_0}^{I_0}$, $T_{C_n}^{C_0}$, $T_{I_n}^{C_n}$, and $T_{I_n}^{I_0}$ represent the transformation matrix from Camera to IMU at t0, Camera at tn to Camera at t0, IMU at tn to camera at tn, and IMU at tn to IMU at t0.

Then, construct the critical online IMU/camera calibration system with the initial rotation. The state vector is set as:

$$X_{IC} = [x_0 \quad \ldots \quad x_m \quad s_0 \quad \ldots \quad s_n \quad q_I^C \quad p_I^C] \quad (2)$$

Where, $x_k$ includes the pose and velocity of the vehicle at time k and the bias of acceleration and gyroscope. $s_0$ represents the inverse depth of the n-th feature points in the environment. $q_I^C$ is the rotation quaternion from IMU to Camera and $p_I^C$ is the translation from IMU to Camera. $T_{I_n}^{C_n}$ is composed of $q_I^C$ and $p_I^C$.

The optimization objective function of the tight-integrated online calibration IMU/camera system is set according to VINS-MONO[29]:

$$f(\xi_{CI}) = argmin \{\|r_p - J_p X\|^2 + \|\gamma_{IMU}(z,X)\|^2 + \|\gamma_C(z,X)\|^2\} \quad (3)$$

Where, the three error items are residual of marginalization, IMU measurement and reprojection error of camera, respectively. Especially, the residual of the camera is composed of the extrinsic parameter of IMU and camera. By optimizing the objective function in Eq.3, the extrinsic of IMU/camera and the pose of mobility vehicle can be estimated online and real time. In addition the initialization of gravity and velocity is also introduced as the same as VINS-MONO.

### B. IMU/Lidar Calibration

To ensure the fast convergence of the online calibration of IMU/Lidar, the hand-eye calibration method is applied to roughly estimate the rotation of IMU and Lidar:

$$T_{L_0}^{I_0} T_{L_n}^{L_0} T_{I_n}^{L_n} = T_{I_n}^{I_0} \quad (4)$$

Where, $T_{L_0}^{I_0}$, $T_{L_n}^{L_0}$, $T_{I_n}^{L_n}$, and $T_{I_n}^{I_0}$ represent the transformation matrix from Lidar to IMU at t0, Lidar at tn to Camera at t0, IMU at tn to Lidar at tn, and IMU at tn to IMU at t0. With the estimated rotation of IMU/Lidar, we can undistort the Lidar points by the large motion.

Then, construct the critical online IMU/Lidar calibration system with the initial rotation. The state vector is set as:

$$X_{IL} = [x_0 \quad \ldots \quad x_m \quad q_I^L \quad p_I^L] \quad (5)$$

Where, $x_k$ includes the pose and velocity of the vehicle at time k and the bias of acceleration and gyroscope. $q_I^L$ is the rotation quaternion from IMU to Lidar and $p_I^L$ is the translation from IMU to Lidar. $T_{I_n}^{L_n}$ is composed of $q_I^L$ and $p_I^L$.

The optimization objective function is set according to LIO-SAM[30]:

$$f(\xi_{LI}) = argmin\{\|\gamma_{IMU}(z,X)\|^2 + \|\gamma_{Plane}(z,X)\|^2 + \|\gamma_{Line}(z,X)\|^2\} \quad (6)$$

Where, the three error items are residual of IMU measurement, the distance of Lidar plane point to map and distance of Lidar line point to map, respectively. Especially, the residual of plane and line point between Lidar and map is composed of the external parameter of IMU and Lidar. By optimizing the objective function in Eq.3, the extrinsic of IMU/Lidar and the pose of mobility vehicle can be estimated online and real time. In addition, the initialization result of gravity and velocity in Sec III-C is introduced for the initial vale of Lidar/IMU online calibration.

### C. Camera/Lidar Association

In order to add the constrain of camera and Lidar, the line and plane feature points-based camera/lidar association is proposed. At first, calculate the initial transformation matrix from Lidar to camera with the estimated matrix of IMU/camera and IMU/Lidar:

$$T_I^C T_L^I = T_L^C \quad (7)$$

Where, $T_I^C$, $T_L^I$, and $T_L^C$ represents the transformation matrix from IMU to camera, Lidar to IMU, and Lidar to camera, respectively.

Then, extract the 3D line and plane feature points in the chessboard from image and Lidar. The introduce of the line feature points can improve the observability of the calibration system. Finally, we can translate the feature points from Lidar to camera and find the associate of the Lidar and camera points by the nearest principle. Fig.3 shows the extracted points from Lidar and camera.

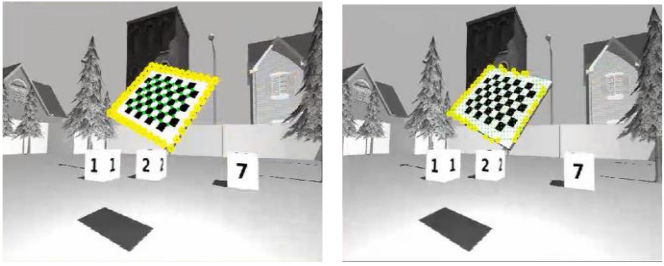

Fig 3. The extracted points from Lidar and camera.

### D. IMU/Camera/Lidar Calibration

We can construct the constrain problem of Lidar and camera with the chessboard in Fig 3. The state vector is set as:

$$X_{ICL} = [x_0 \quad \dots \quad x_m \quad q_I^C \quad p_I^C \quad q_I^L \quad p_I^L] \quad (8)$$

Where, $x_k$ includes the pose and velocity of the vehicle at time k and the bias of acceleration and gyroscope. $q_I^C$ is the rotation quaternion from IMU to Camera and $p_I^C$ is the translation from IMU to Camera. $T_{I_n}^{C_n}$ is composed of $q_I^C$ and $p_I^C$. $q_I^L$ is the rotation quaternion from IMU to Lidar and $p_I^L$ is the translation from IMU to Lidar. $T_{I_n}^{L_n}$ is composed of $q_I^L$ and $p_I^L$.

The optimization objective function is:

$$f(\xi_{ICL}) = argmin\left\{\|\gamma_{IMU}(z,X)\|^2 + \|\gamma_{chessboard\_L}(z,X)\|^2\right\} \quad (9)$$

Where, the three error items are residual of IMU measurement, distance of lidar plane point to map and distance of lidar line point to map, respectively. Especially, the residual of plane and line point between Lidar and map is composed of the extrinsic parameter of IMU and Lidar. By optimizing the objective function in Eq.9, the extrinsic of IMU/camera, IMU/Lidar and Lidar/camera can be estimated online and real time. The detail of the residual of the chessboard is:

$$\gamma_L\left(\gamma_{chessboard\_L}(z,X)\right) = argmin\left\{\|q_I^C q_L^I P_{Line}^L - P_{Line}^C\|^2 + \|q_I^C q_L^I P_{Plane}^L - P_{Plane}^C\|^2\right\} \quad (10)$$

Finally, with the refined calibration result we can undistort Lidar points continually to improve the location and calibration precision.

## IV. EXPERIMENTS

We conduct the simulation test to show the accuracy of the proposed IMU/camera/Lidar co-calibration method with Lidar VLP-16, MEMS-IMU and camera in a simulated quadrotor. The parameters of the sensors are set as Table I. To calibrate the 6D pose for the three sensors, the 3D rotation and translation movements is necessary and the chessboard is arranged in the environment as shown in Fig.4.

TABLE I. SENSOR PARAMETERS

| Sensors | Parameter | Frequency |
|---|---|---|
| Accelerometer noise density | 60 μg/$\sqrt{Hz}$ | 400Hz |
| Accelerometer in-run bias stability | 15 μg | 400Hz |
| Gyroscope noise density | 0.01°/s/$\sqrt{Hz}$ | 400Hz |
| Gyroscope bias | 10 deg/h | 400Hz |
| Lidar noise | 0.03m | 10Hz |
| Camera Pixel Size | 2μm × 2μm | 50Hz |

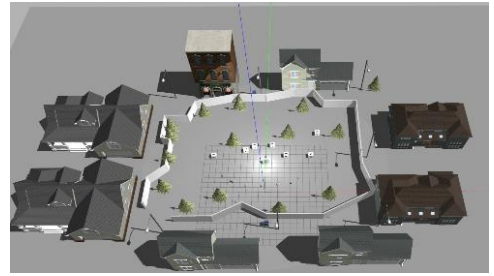

Fig 4. Simulation environment.

The proposed method is compared with the two-step method. In the two-step method, we calibrate the IMU and camera with kalibr and then calibrate IMU and Lidar with the open source method in [8], which is modified to only one iteration for real test. Afterwards, the translation matrix from Lidar to camera can been calculated. Finally, project the Lidar point to the image to show the efficiency of the proposed method, as shown in Fig.5. The calibration result of IMU/Lidar and IMU/camera are shown in Table II and we can see that the accuracy of the proposed method is improved greatly.

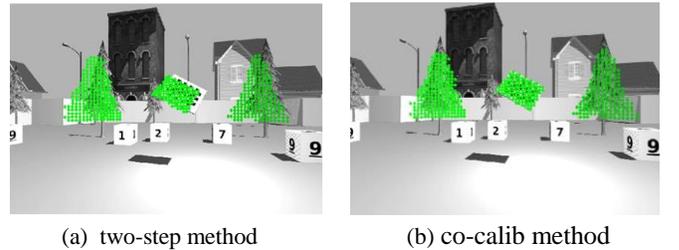

(a) two-step method     (b) co-calib method

Fig 5. The results of Lidar/camera calibration of the two-step method and the proposed co-calib method.

TABLE II. CALIBRATION RESULTS

| Sensors | | Calibration value | | |
|---|---|---|---|---|
| | | *Ture value* | *Two-step* | *Co-calib* |
| IMU/Lidar | Rotation_x(rad) | 0.16 | 0.172 | 0.164 |
| | Rotation_y(rad) | 0.16 | 0.167 | 0.158 |
| | Rotation_z(rad) | 0.16 | 0.165 | 0.158 |
| | Tranlation_x (m) | 0.1 | 0.109 | 0.102 |

| Sensors | | Calibration value | | |
|---|---|---|---|---|
| | | *Ture value* | *Two-step* | *Co-calib* |
| IMU/Camera | Tranlation_y (m) | 0 | 0.017 | 0.001 |
| | Tranlation_z (m) | -0.08 | -0.071 | -0.081 |
| | Rotation_x (rad) | -0.16 | -0.165 | -0.165 |
| | Rotation_y (rad) | -0.16 | -0.159 | -0.161 |
| | Rotation_z (rad) | -0.04 | -0.046 | -0.044 |
| | Tranlation_x (m) | 0.1 | 0.123 | 0.103 |
| | Tranlation_y (m) | 0 | 0.011 | 0.001 |
| | Tranlation_z (m) | 0 | 0.023 | 0.003 |

## V. CONCLUSION

To improve the accuracy of IMU/Lidar/Camera calibration, an optimization-based Co-calibration method is proposed in the paper. To fuse the mutual constraints of IMU/camera, IMU/Lidar and Lidar/camera, the IMU/camera and IMU/Lidar online calibration is conducted first. And then the corner and surface feature points in the chessboard are associated with the coarse calibration result of IMU/camera and IMU/Lidar to construct the camera/Lidar constraint. Finally, the co-calibration optimization is constructed to refine all extrinsic parameters. In the simulation test, it is demonstrated that our proposed method outperforms the normal two-step method.

## ACKNOWLEDGMENT

This work was supported by the National Natural Science Foundation of China under Grant 62073080 and Grant 61921004, the Fundamental Research Funds for the Central Universities under Grant 2242022K30017 and Grant 2242022K30018, and China Scholarship Council.

## REFERENCES

[1] Shan T, Englot B, Ratti C et al. LVI-SAM: Tightly-coupled Lidar-Visual-Inertial Odometry via Smoothing and Mapping[C]. 2021 IEEE international conference on robotics and automation (ICRA). IEEE, 2021: 5692-5698.

[2] Yin J, Li A, Li T et al. M2DGR: A Multi-Sensor and Multi-Scenario SLAM Dataset for Ground Robots[J]. IEEE Robotics and Automation Letters, 2022, 7(2): 2266-2273.

[3] Ou J, Huang P, Zhou J et al. Automatic Extrinsic Calibration of 3D LIDAR and Multi-Cameras Based on Graph Optimization[J]. Sensors, 2022, 22(6): 2221.

[4] Mirzaei F M R S. A Kalman Filter-based Algorithm for IMU-Camera Calibration[J]. IEEE transactions on robotics, 2008, 5(24): 1143-1156.

[5] Rehder J, Nikolic J, Schneider T et al.Extending kalibr: Calibrating the extrinsics of multiple IMUs and of individual axes[C]. 2016 IEEE International Conference on Robotics and Automation (ICRA). IEEE, 2016: 4304-4311.

[6] Xiao X, Zhang Y, Li H et al. Camera-IMU Extrinsic Calibration Quality Monitoring for Autonomous Ground Vehicles[J]. IEEE Robotics and Automation Letters, 2022, 7(2): 4614-4621.

[7] Lv J, Xu J, Hu K et al. Targetless Calibration of LiDAR-IMU System Based on Continuous-time Batch Estimation[C]. 2020 IEEE/RSJ International Conference on Intelligent Robots and Systems (IROS). IEEE, 2020: 9968-9975.

[8] Lv J, Zuo X, Hu K et al. Observability-Aware Intrinsic and Extrinsic Calibration of LiDAR-IMU Systems[J]. IEEE Transactions on Robotics, 2022: 1-20.

[9] Le Gentil C, Vidal-Calleja T, Huang S. 3D Lidar-IMU Calibration based on Upsampled Preintegrated Measurements for Motion Distortion Correction[C]. 2018 IEEE International Conference on Robotics and Automation (ICRA). IEEE, 2018: 2149-2155.

[10] Liu W, Li Z, Malekian R et al. A Novel Multifeature Based On-Site Calibration Method for LiDAR-IMU System[J]. IEEE Transactions on Industrial Electronics, 2020, 67(11): 9851-9861.

[11] Zuo X G P L. LIC-Fusion: LiDAR-Inertial-Camera Odometry[C]. 2019 IEEE/RSJ International Conference on Intelligent Robots and Systems (IROS). IEEE, 2019: 5848-5854.

[12] Zuo X, Yang Y, Geneva P et al. LIC-Fusion 2.0: LiDAR-Inertial-Camera Odometry with Sliding-Window Plane-Feature Tracking[C]. 2020 IEEE/RSJ International Conference on Intelligent Robots and Systems (IROS). IEEE, 2020: 5112-5119.

[13] Verma S B J S W. Automatic extrinsic calibration between a camera and a 3D Lidar using 3D point and plane correspondences[C]. 2019 IEEE Intelligent Transportation Systems Conference (ITSC). IEEE, 2019: 3906-3912.

[14] Kümmerle J K T. Unified Intrinsic and Extrinsic Camera and LiDAR Calibration under Uncertainties[C]. 2020 IEEE International Conference on Robotics and Automation (ICRA). IEEE, 2020: 6028-6034.

[15] Kim E, Park S. Extrinsic Calibration between Camera and LiDAR Sensors by Matching Multiple 3D Planes[J]. Sensors, 2020, 20(1): 52.

[16] Xie Y, Deng L, Sun T et al. A4LidarTag: Depth-Based Fiducial Marker for Extrinsic Calibration of Solid-State Lidar and Camera[J]. IEEE robotics and automation letters, 2022, 7(3): 6487-6494.

[17] Zhen W, Hu Y, Liu J et al. A Joint Optimization Approach of LiDAR-Camera Fusion for Accurate Dense 3-D Reconstructions[J]. IEEE Robotics and Automation Letters, 2019, 4(4): 3585-3592.

[18] Nagy B, Kovacs L, Benedek C.Online Targetless End-to-End Camera-LIDAR Self-calibration[C]. 2019 16th International Conference on Machine Vision Applications (MVA), 2019: 1-6.

[19] Yuan C L X H. Pixel-level Extrinsic Self Calibration of High Resolution LiDAR and Camera in Targetless Environments[J]. IEEE Robotics and Automation Letters, 2021, 6(4), 7517-7524..

[20] Ravi R, Lin Y, Elbahnasawy M et al. Simultaneous System Calibration of a Multi-LiDAR Multicamera Mobile Mapping Platform[J]. IEEE Journal of Selected Topics in Applied Earth Observations and Remote Sensing, 2018, 11(5): 1694-1714.

[21] Kim D, Shin S, Kweon I S. On-Line Initialization and Extrinsic Calibration of an Inertial Navigation System With a Relative Preintegration Method on Manifold[J]. IEEE Transactions on Automation Science and Engineering, 2018, 15(3): 1272-1285.

[22] Pusztai Z, Eichhardt I, Hajder L. Accurate Calibration of Multi-LiDAR-Multi-Camera Systems[J]. Sensors, 2018, 18(7): 2139.

[23] Diehm A L, Gehrung J, Hebel M et al.Extrinsic self-calibration of an operational mobile LiDAR system[Z]. Laser Radar Technology and Applications, 11410: 46-61.

[24] Tschopp F, Riner M, Fehr M et al. VersaVIS—An Open Versatile Multi-Camera Visual-Inertial Sensor Suite[J]. Sensors, 2020, 20(5): 1439.

[25] Chuan F. Single-Shot is Enough: Panoramic Infrastructure Based Calibration of Multiple Cameras and 3D LiDARs[C]. 2021 IEEE/RSJ International Conference on Intelligent Robots and Systems (IROS). IEEE, 2021: 8890-8897.

[26] J. D, J. F P K, D. M G.An Extrinsic Calibration Tool for Radar, Camera and Lidar[Z]. 2019 International Conference on Robotics and Automation (ICRA). IEEE, 2019: 8107-8113.

[27] Pentek Q, Kennel P, Allouis T et al. A flexible targetless LiDAR–GNSS/INS–camera calibration method for UAV platforms[J]. ISPRS Journal of Photogrammetry and Remote Sensing, 2020, 166: 294-307.

[28] Extrinsic Calibration of LiDAR, IMU and Camera[J]. 2022. arXiv preprint arXiv:2205.08701.

[29] Qin, T., & Shen, S.. Online temporal calibration for monocular visual-inertial systems[C]. 2018 IEEE/RSJ International Conference on Intelligent Robots and Systems (IROS). IEEE, 2018:3662-3669.

[30] Shan, T., Englot, B., Meyers, D., Wang, W., Ratti, C., & Rus, D. Lio-sam: Tightly-coupled lidar inertial odometry via smoothing and mapping[C]. 2020 IEEE/RSJ international conference on intelligent robots and systems (IROS) . IEEE, 2020:5135-5142.